\documentclass[10pt,twocolumn,letterpaper]{article}

\usepackage[pagenumbers]{cvpr} %

\usepackage{enumitem}
\usepackage{comment}
\usepackage{graphicx}
\usepackage{multirow}
\usepackage{amsmath}
\usepackage{dsfont}
\usepackage{lipsum}
\usepackage{caption}

\usepackage{pifont}
\usepackage{amssymb}
\newcommand{\cmark}{\ding{51}}       %
\newcommand{\xmark}{\ding{55}}       %
\newcommand{\pmark}{$\blacktriangle$} %

\usepackage{colortbl}
\usepackage{subcaption}

\newif\ifshowTODOs
\showTODOsfalse %

\usepackage{microtype}

\renewcommand{\paragraph}[1]{\vspace{.5em}\noindent\textbf{#1.}}

\usepackage{soul}
\setuldepth{foobar}

\newcommand{\DATAP}{\emph{MisEngine}}
\newcommand{\MODEL}{\emph{MisFormer}}
\definecolor{cvprblue}{rgb}{0.21,0.49,0.74}
\usepackage[pagebackref,breaklinks,colorlinks,allcolors=cvprblue]{hyperref}

\def\paperID{13225} %
\def\confName{CVPR}
\def\confYear{2026}

\title{Mistake Attribution: Fine-Grained Mistake Understanding in Egocentric Videos}

\author{
    Yayuan Li$^{1}$\quad Aadit Jain$^{1}$\quad Filippos Bellos$^{1}$\quad Jason J. Corso$^{1,2}$ \\
    $^{1}$University of Michigan \quad $^{2}$Voxel51
    \\
    \small\href{https://yayuanli.github.io/MATT}{\texttt{https://yayuanli.github.io/MATT}}
  }

\begin{document}
\maketitle

\begin{abstract}
We introduce Mistake Attribution (MATT), a new task for fine-grained understanding of human mistakes in egocentric videos.
While prior work detects whether a mistake occurs, MATT attributes the mistake to \emph{what} part of the instruction is violated (semantic role), \emph{when} in the video the deviation becomes irreversible (the Point-of-No-Return, PNR), and \emph{where} the mistake appears in the PNR frame.
We develop MisEngine, a data engine that automatically constructs mistake samples from existing datasets with attribution-rich annotations.
Applied to large egocentric corpora, MisEngine yields EPIC-KITCHENS-M and Ego4D-M---two datasets up to two orders of magnitude larger than prior mistake datasets.
We then present MisFormer, a unified attention-based model for mistake attribution across semantic, temporal, and spatial dimensions, trained with MisEngine supervision.
A human study demonstrates the ecological validity of our MisEngine-constructed mistake samples, confirming that EPIC-KITCHENS-M and Ego4D-M can serve as reliable benchmarks for mistake understanding.
Experiments on both our datasets and prior benchmarks show that MisFormer, as a single unified model, outperforms task-specific SOTA methods by at least 6.66\%, 21.81\%, 18.7\%, and 3.00\% in video-language understanding, temporal localization, hand-object interaction, and mistake detection, respectively.

\end{abstract}
  
\section{Introduction}
\label{sec:intro}

\begin{figure}[t]
    \centering
        \includegraphics[width=1\linewidth]{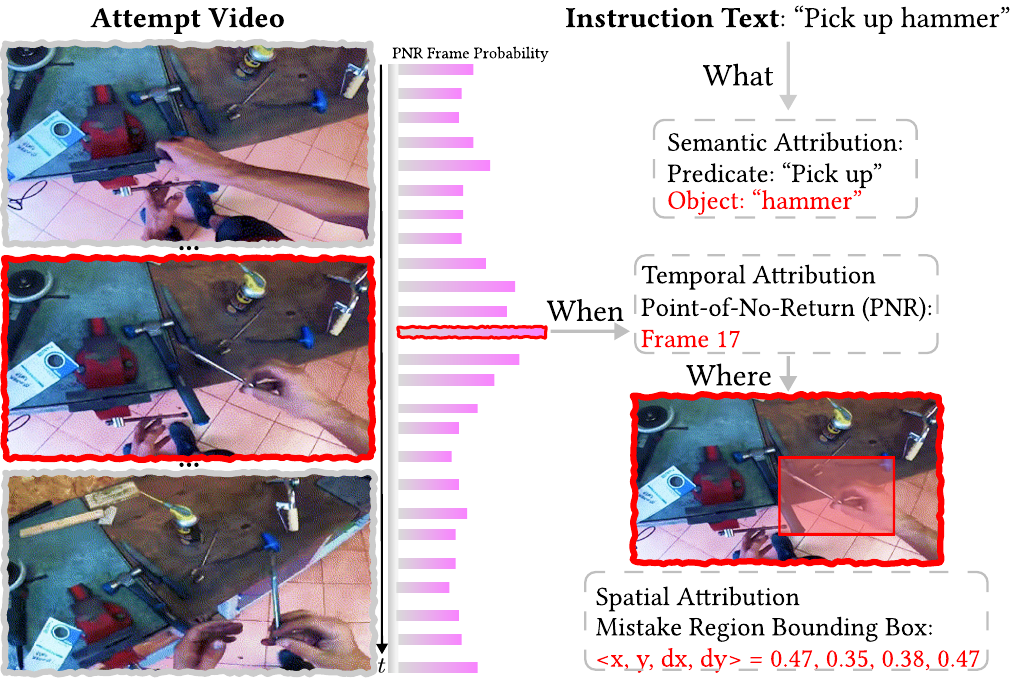}
    \caption{
        Mistake Attribution (MATT) task aims to understand the deviation between a human attempt (video) and the instruction (text) along three axes. Semantic attribution identifies \textbf{what} semantic role in the instruction is violated (e.g., a wrong Object ``bolt'' is mistakenly picked up instead of ``hammer''); temporal attribution identifies \textbf{when} the attempt reaches the point of no return (PNR) (e.g., Frame 17); and spatial attribution identifies \textbf{where}, in the PNR frame, the mistake is manifested (e.g., the red bounding box).
    }
    \label{fig:intro}
\end{figure}
Physically-grounded instructional AI assistants that provide guidance and feedback given egocentric video has drawn significant attention in the vision community~\cite{Damen2022RESCALING,wang2023holoassistegocentrichumaninteraction,slawson2018case,bellos2025towards}.
Leveraging clear hand and object interaction~\cite{singh2024handobjectinteractionpretrainingvideos,Kwon_2021_ICCV} along with rich task models~\cite{10.1109/TPAMI.2019.2927476}, these assistants show significant promise in the way humans acquire skills~\cite{EPPLER1995391,slawson2018case} across various physical activities, spanning everyday activities like cooking, cleaning, and maintenance~\cite{youcook2,grauman2022ego4d, Damen2022RESCALING,sener2022assembly101,miech2019howto100m, wang2023holoassistegocentrichumaninteraction} to  professional tasks~\cite{haque2020illuminating, garrow2021machine, zisimopoulos2018deepphase, ozsoy2025egoexor, DEVAGIRI2022118002}.

A critical aspect of good physically-grounded AI assistance is the ability to steer fallible (all) humans away from mistakes, like using salt instead of sugar or applying the wrong type of bandage onto a wound.  Human mistakes result from distractions, misunderstanding, or ambiguities in task formulation~\cite{sener2022assembly101}, even when given detailed instructions. 
Given their importance, handling mistakes in egocentric video has seen increasing attention in recent years.  
Most works focus on detecting step-level mistakes like missing or erroneous actions~\cite{soran2015generating, wang2023holoassistegocentrichumaninteraction,flaborea2024prego} often modeled via anomalies~\cite{lee2024error,mazzamuto2025gazing} or classifying observed mistakes, e.g., step modification and slip errors~\cite{lee2025error}. More recently, MistScene~\cite{patsch2025mistsense} generates natural language explanations for an action video that is detected as a mistake. 

Modeling mistakes more finely is integral to realizing the potential of AI assistants.  
For example, the coarse categorization in \cite{soran2015generating, wang2023holoassistegocentrichumaninteraction,flaborea2024prego,peddi2024captaincook4d} does not infer what specific component of the activity is a mistake.  Similarly, the natural language explanation of MistScene~\cite{patsch2025mistsense}, while impressive, focus only on inherent activity mistakes with no alignment to the actual task instructions. 
Specifically, current works cannot answer key questions such as \textbf{what} part of the instruction is not followed, \textbf{when} during the user's action is the mistake's impact irreversible (i.e., the Point-of-No-Return), and \textbf{where} in the PNR frame does the mistake manifest. 
For example, in Fig.~1, the instruction is ``pick up hammer," but the executed action in the video is ``pick up bolt." Mistake Detection can reveal that a mistake has occurred, but does not show that the expected object ``hammer" is not followed (\textbf{what}), \textbf{when} the mistake happens during execution, or the region in the red box \textbf{where} the mistake manifested.

To fill this gap, we introduce Mistake Attribution (MATT), which focuses on fine-grained understanding of deviations between the instruction and the observed egocentric user's action. Specifically, mistake attribution consists of three components: (1) semantic attribution, which seeks to assess \emph{what}---i.e., which semantic role\footnotemark[1] from the instruction---is mistaken in the attempt video; (2) temporal attribution, which identifies \emph{when} the mistake is consolidated by pinpointing the point-of-no-return (PNR) frame in the attempt video~\cite{grauman2022ego4d}; and (3) spatial attribution, which determines \emph{where} the mistake occurs within a precise region of the PNR frame. 
MATT 
provides a triplet of semantic-temporal-spatial information about the mistake, which provides rich information for AI Assistants along with other performance applications like self-learning.

\footnotetext[1]{Semantic roles refer to the underlying relationships that components in a sentence have with the main activity, such as predicate (the main verb of the activity), object (the noun the predicate is acting on), or instrument (the tool the predicate is using), as discussed in Fillmore (1968) \cite{fillmore1968case}.}

We introduce two vectors of contributions for this new mistake attribution problem.  First, we propose a novel data engine, MisEngine, that automatically constructs  large-scale mistake samples from existing action-recognition sources.  Existing mistake datasets lack triplet-attribution annotations (for training and evaluation) and are limited in quantity and diversity---making them insufficient to benchmark MATT at real-world scale.  Importantly, collecting such a dataset is non-trivial: as collectors gain experience, mistakes become rarer and manual collection grows inefficient, while, at the same time, injecting staged mistakes into datasets introduces visual bias that pushes the dataset away from what one expects in the real world.  Consequently, existing datasets~\cite{lee2024error,sener2022assembly101,ding2023every,wang2023holoassistegocentrichumaninteraction,peddi2024captaincook4d,jang2019epic} are often orders of magnitude smaller and less diverse than general action-understanding corpora (\cref{tab:dataset-comparison}).

To overcome these bottlenecks, MisEngine systematically cross-matches an instruction text with descriptions of other action videos to create misaligned pairs of instruction text and action video. It leverages Semantic Role Labeling (SRL) to control cross-matching by semantic groups (e.g., Predicate, Object)---inherently yielding semantic-attribution labels---while videos naturally inherit temporal and spatial annotations that we process into the corresponding attribution targets.

Second, we introduce a transformer-based model, MisFormer, that is able to address MATT's breadth in a single, unified model.   MisFormer comprises a dual-branch feature extractor (video and text) with cross-attention and three modules that consume role-encoded instruction tokens to produce semantic, temporal, and spatial attributions. 
For semantic attribution, a decoder treats the instruction representation as queries and video features as keys/values---yielding \{Correct, Mistake\} for each semantic role. 
For temporal attribution, the model downsamples video features and uses another encoder, cross-attending to the instruction, to localize the PNR frame. 
For spatial attribution, it derives a bounding box from attention between the PNR frame and the instruction representation; at inference, the temporal and spatial modules are gated---invoked only when any role is predicted as Mistake.

We evaluate our work on two popular egocentric action datasets—Ego4D~\cite{grauman2022ego4d} and EPIC-KITCHENS~\cite{Damen2022RESCALING}---covering diverse real-world activities. Applying MisEngine yields Ego4D-M and EPIC-KITCHENS-M, the first datasets for training and benchmarking MATT and mistake understanding at real-world scale. We evaluate MisFormer against strong baselines, including Video–Language Models \cite{openai2024gpt4o,maaz-etal-2024-video,zhang2024llavanextvideo}, Temporal Localization Models \cite{lei2022masked,xue2023egot2}, Hand–Object Interaction detectors \cite{mediapipe,leonardi2024synthetic}, and mistake detection methods~\cite{AMNAR_Huang_2025_CVPR,lee2024error}, on both our and prior benchmarks. Although each baseline focuses on a single attribution task, they underperform on EPIC-KITCHENS-M and Ego4D-M, highlighting the difficulty of MATT. MisFormer, as a unified model, achieves superior performance or efficiency across tasks.

\section{Related Work}

\noindent\textbf{Modeling Mistakes in Egocentric Videos.}\quad
Recent efforts toward Instructional AI have demonstrated early progress~\cite{bellos2025towards,li2024handi,storks2025transparent,li2025towards, bellos-etal-2024-large,wang2026bimotion,bi2026think}.
In terms of Misatake Understanding, although there is existing work in video anomaly detection~\cite{NAYAK2021104078,zhaounusual2011,anomalynet2019,10.1007/978-3-031-19830-4_23,hasan2016learning}, most of this work does not emphasize anomalies as mistakes in the context of a procedure.
Most past mistake-specific work focuses on detection or small-set categorization.   Soran et al.~\cite{soran2015generating} identify omitted actions using a hidden Markov models.
HoloAssist~\cite{wang2023holoassistegocentrichumaninteraction} uses the TimeSFormer ViT model \cite{bertasius2021space,dosovitskiy2021an} to detect mistakes.
Assembly101~\cite{sener2022assembly101, ding2023every} uses a graph-based method to detect two categories of mistakes specifically in the assembly task: the ordering and placement of installing a component.
Certain recent works also adopt the anomaly detection mindset for mistakes, such as EgoPED~\cite{lee2024error}, which compares an observed action feature to its prototypical one,  PREGO~\cite{flaborea2024prego}, which compares the inferred action class to the expected one in a streaming manner, AMNAR~\cite{AMNAR_Huang_2025_CVPR}, which combines these two ideas, and Missteps~\cite{mazzamuto2025gazing}, which compares observed gaze behavior against an expected one.  None of these works models mistakes at a level of richness beyond binary detection. 

The most relevant recent works are twofold. 
Lee et al.~\cite{lee2025error} categorize detected mistakes into a set of predefined categories (e.g., modification, slip errors, etc.). 
MistSense~\cite{patsch2025mistsense,storks2025transparent} generates natural language explanations for an action video that is detected as mistaken.
Neither of these works captures a sufficiently rich task-based analysis of the mistake.  The categories in Lee et al.~\cite{lee2025error} do not describe the specific attributes of the mistake.  MistSense's~\cite{patsch2025mistsense} explanation capture generic action failures (e.g., knocking over the bottle) rather than how the action deviates from the task instructions.

\paragraph{Datasets for Egocentric Mistake Understanding}
Existing mistake understanding datasets (EgoPER~\cite{lee2024error}, Assembly101~\cite{sener2022assembly101, ding2023every}, HoloAssist~\cite{wang2023holoassistegocentrichumaninteraction}, CaptainCook4D~\cite{peddi2024captaincook4d}, Epic-Tent~\cite{jang2019epic}) are small and exhibit limited semantic and visual variance, hindering real‑world benchmarking (\cref{tab:dataset-comparison} quantifies these).  Most provide mistake detection and recognition labels but insufficient supervision for MATT attribution. Where mistake explanations exist, they are free‑form rather than structurally aligned to instruction semantic roles, impeding automation. Precise error timing and grounded regions are typically missing; point‑of‑no‑return timestamps and mistake‑grounding boxes are absent or only indirectly inferable. 
Conversely, large egocentric video datasets such as EPIC‑KITCHENS~\cite{Damen2022RESCALING} and Ego4D~\cite{grauman2022ego4d} offer scale and diversity for action understanding.  Yet, they lack native mistake instances and structured attribution, and are hence not directly suitable for mistake benchmarking.  In this paper, our MisEngine combines the benefits of the two types of datasets, constructing benchmarks with annotations of semantic, temporal and spatial attribution at scale at least two orders of magnitude larger than existing mistake datasets.

\section{Mistake Attribution}
\label{sec:problem}
Consider a sequence of activities comprising a goal oriented task, such as cooking a dish or assembling a toy.  At any moment when carrying out the task, it is necessary to perform a particular action in a particular way. We assume these actions are adequately described in the instruction text.  Work in mistake understanding generally seeks to capture when, for one of these task-based actions, the way in which the user performs the action deviates from the instruction text.  These deviations can include performing the wrong action entirely, using the wrong tool, or working with the wrong object.  

Mistake Attribution (MATT) seeks to model these various fine-grained types of deviations.  Concretely, given an instruction text $T$ (e.g., ``cut the apple'') and an egocentric video $V$ of the user attempting to perform $T$, which we call an \textit{attempt video},  MATT is composed of the following:

\noindent \textbf{Semantic Attribution}\, identifies the specific components (i.e., semantic roles~\cite{Fillmore1968}) in the instruction text $T$ that are not correctly followed in the attempt video $V$. 
Formally, we define the semantic attribution output as ${\{y_r \in \{0,1\} \mid r \in \mathcal{R}\}}$, denoting whether the semantic role $r$ was correctly followed ($y_r = 0$) or no ($y_r = 1$) in the attempt video. $\mathcal{R}$ denotes the set of semantic roles in the instruction text. 

\noindent\textbf{Temporal Attribution}\, specifies the Point-of-No-Return~(PNR)~\cite{grauman2022ego4d} in the attempt video $V$ after which where the impact of the mistake is irreversible. We denote this PNR timestamp as $t_{\mathrm{PNR}} \in \mathcal{T}$ where $\mathcal{T}$ is the total number of frames in the attempt video. 

\noindent\textbf{Spatial Attribution}\, highlights the spatial region in the PNR frame to indicate the visual details of the mistake. Specifically, it produces the mistake grounding box $B_{t_{\mathrm{PNR}}} = (x_{\min}, y_{\min}, \delta_{x}, \delta_{y})$, localizing the region that suggests the mistake within the PNR timestamp $t_{\mathrm{PNR}}$.
To address MATT, a function $\mathcal{F}$ needs to map the input pair of instruction text and attempt video to the triplet of semantic-temporal-spatial mistake attribution:
\begin{align}
    \langle (y_r)_{r \in \mathcal{SR}},\ t_{\mathrm{PNR}},\ B_{t_{\mathrm{PNR}}} \rangle
    =
    \mathcal{F}(T, V) .
\end{align}
\noindent In the two sections that follow, we describe the data engine for creating a large scale mistake dataset for MATT and a unified model capable of addressing MATT.

\subsection{MisEngine: Automated large-scale mistake dataset construction}
\label{sec:dataset-construction}
\begin{figure*}[t!]
    \centering
    \includegraphics[width=1\textwidth]{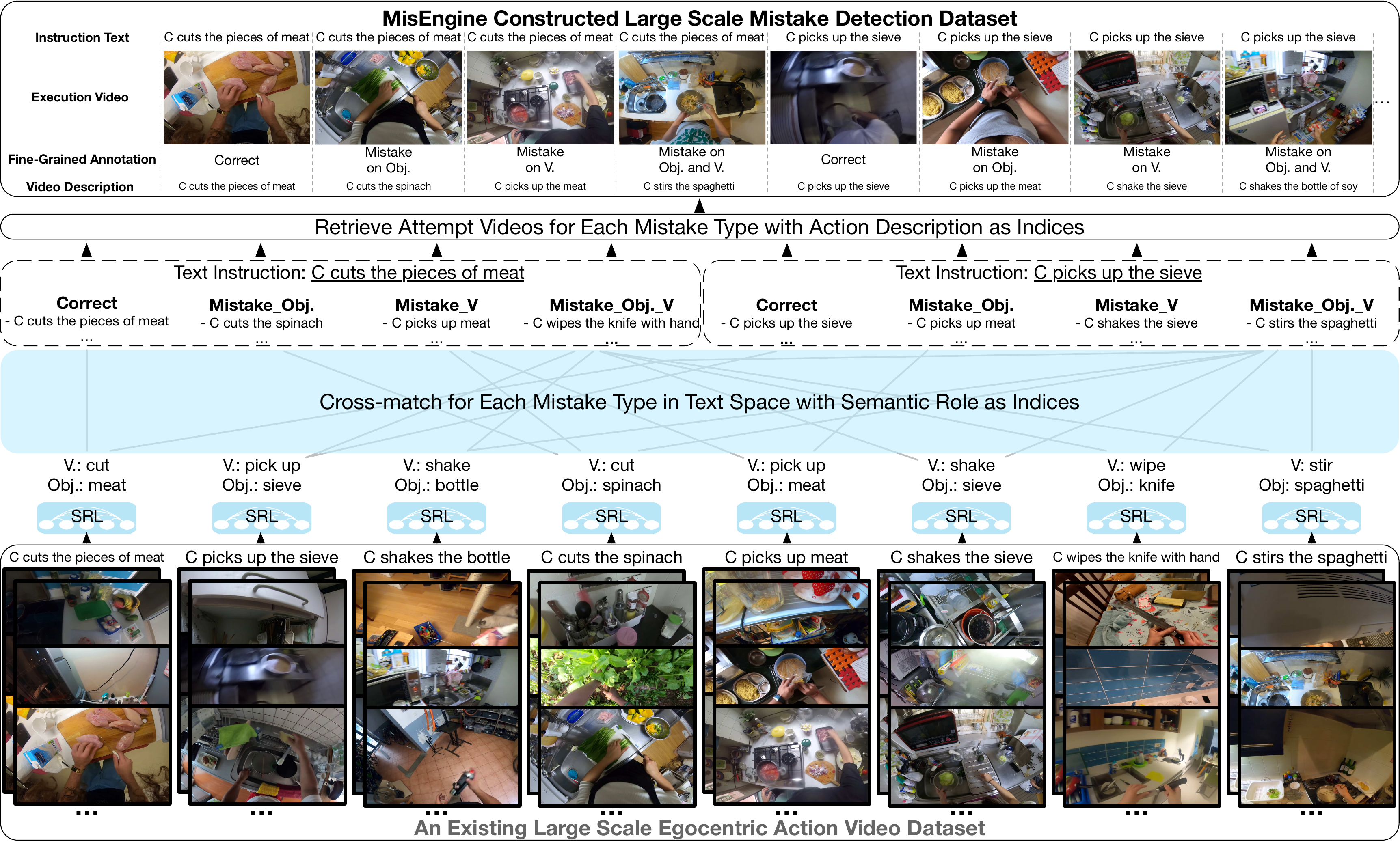}
    \caption{A significant challenge in mistake video analysis is the paucity of available data in the face is massive diversity of possible mistakes.  This figure explains our new data engine, MisEngine, that overcomes this challenge by automatically creating new mistake understanding datasets from source corpora by a careful series of sampling and cross-matching methods.  MisEngine uses semantic role labeling on the text instruction and then matches across the available roles; here, we show an example with two roles (object as ``Obj'' and predicate as ``V'').  Our resulting datasets are orders of magnitude larger than existing ones and fully annotated (for free) across mistake detection and attribution. 
    }
    \label{fig:architecture_dataset}
\end{figure*}

\paragraph{Mistake Samples Construction}
A source action recognition dataset $\Lambda$ contains pairs of action descriptions and action videos. MisEngine follows three steps to construct the mistake dataset $\Omega$ from $\Lambda$.

Step 1: We apply Semantic Role Labeling (SRL)~\cite{Fillmore1968,allen_srl} to parse each action description (instruction) text in $\Lambda$. For the $i$th action description, SRL produces a set $G_i = \{g^r_i \mid r \in \mathcal{R}\}$ containing all semantic groups for this action description where $\mathcal{R}$ is the set of all semantic roles defined by the SRL method. 
As illustrated in~\cref{fig:architecture_dataset} with $\mathcal{R} = \{predicate, object\}$ as an example, the action description text ``Pick up the sieve'' is parsed into $G_i = \{g^{Predicate}_i=\text{``Pick up''}$, $g^{Object}_i=\text{``the sieve''}\}$. 

Step 2: For each $G_i$, we compare with all other parsed actions descriptions $G_j$ ($j \neq i$) by checking with a function $\mathcal{J}$ if the semantic group $g^r_i$ and $g^r_j$ are the same (i.e., $\mathcal{J}(g^r_i, g^r_j)=1$) for each semantic role $r$. 
$\mathcal{J}$ can be simple character-level matching or semantic-level matching, depending on granularity of the mistake one wants to capture. 
This results in $C = |\mathcal{R}|^2$ misalignment categories for each and all parsed action descriptions $G_i$. 
Taking $\mathcal{R} = \{predicate, object\}$ as an example, the total number of misalignment categories is $C = 4$: Predicate Mistake, Object Mistake, Both Mistakes, and No Mistake. 

Step 3: We sample $N_{T}^{C}$ action descriptions from each of $C$ misalignment categories. For each of these action description, $N_{V}^{c}$ of the videos are sampled as attempt videos for $G_i$. 
For example, if $G_i = \{g^{Predicate}_i=\text{``Pick up''}$, $g^{Object}_i=\text{``the sieve''}\}$ and $G_j = \{g^{Predicate}_j=\text{``Pick up''}$, $g^{Object}_j=\text{``the pan''}\}$, then videos originally associated with the action description ``Pick up the pan'' in $\Lambda$ are randomly sampled (without replacement) as a subset of attempt videos for $G_i$. 
This process results in the number of attempt videos for each $G_i$ as $\Gamma = C \times N_{T}^{C} \times N_{V}^{c}$. 
Denote the number of instruction texts in the constructed dataset $\Omega$ as $N_{T}$. The number of samples (pairs of instruction text and attempt video) in the target dataset is:
\begin{align}
    |\Omega| = N_{T} \times \Gamma = N_{T} \times C \times N_{T}^{C} \times N_{V}^{c}
    \label{eq:dataset_size}
\end{align}

\paragraph{Mistake Attribution Annotation}
The MisEngine workflow induces the semantic attribution annotation $(y_r)_{r \in \mathcal{R}}$ and inherits the temporal attribution annotation $t_{\mathrm{PNR}}$ and spatial attribution annotation $B_{t_{\mathrm{PNR}}}$ from the original action recognition dataset $\Lambda$.   
For semantic attribution, at step 2, the misalignment categories tell if each semantic role is correctly followed or not. 
For temporal attribution, the PNR timestamp in the attempt video $V$ is cloned from the PNR annotation in the original action recognition dataset $\Lambda$. 
For spatial attribution, the mistake grounding box $B_{t_{\mathrm{PNR}}}$ is the unified bounding box from the hand bounding boxes in the PNR frame, annotated in $\Lambda$. 
Hence, MisEngine fully labels the dataset automatically.

There are various action recognition datasets, e.g., Recasens et al.~\cite{NIPS2015_ec895663}, that contains rich information associated with the action videos like eye gaze, head pose, audio, etc. MisEngine enables the possibility of inheriting this information to further study large scale mistake understanding.

\paragraph{MATT Datasets: Ego4D-M and EPIC-KITCHENS-M} We demonstrate MisEngine on two popular action recognition datasets: Ego4D~\cite{grauman2022ego4d} and EPIC-KITCHENS~\cite{Damen2022RESCALING}, resulting in the Ego4D-M and EPIC-KITCHENS-M datasets. These are the first MATT datasets that inherit their quantity and diversity from existing large scale action recognition datasets while labeling the specific attributes necessary for fine-grained mistake modeling and benchmarking. 

For step 1, we use AllenNLP's SRL method~\cite{allen_srl}, which is based on a deep BiLSTM encoder~\cite{he-etal-2017-deep} with attention and a Conditional Random Field~\cite{lafferty2001conditional} output layer to produce the semantic groups for each semantic role in each action description. Since most of the action descriptions in the source datasets contain only the predicate and object, we filter out the shorter descriptions and cut the longer descriptions to keep only the predicate and object, i.e., $\mathcal{R} = \{predicate, object\}$. 
In step 2, we implement the comparing function $\mathcal{J}$ by looking up the original taxonomy annotation in Ego4D or do character-level matching in EPIC-KITCHENS. 
In step 3, we set $N_{T}^{C}$ to 4 and 3 and $N_{V}^{c}$ to 2 and 3 for Ego4D and EPIC-KITCHENS, respectively. We filter out the action descriptions ($G_i$) that have less than $N_{T}^{C}$ misaligned action descriptions for each $C$ or less than $N_{V}^{c}$ candidate videos. 
This results in $N_{T}^{Ego} = 16099$ and $N_{T}^{EK} = 12283$ instruction texts in the Ego4D-M and EPIC-KITCHENS-M datasets, respectively. Following~\cref{eq:dataset_size}, the total number of samples are $|\Omega^{Ego}| = 16099 \times 2 \times 4 \times 2 = 257,584$ and $|\Omega^{EK}| = 12283 \times 2 \times 3 \times 3 = 221,094$ where each has semantic, temporal, and spatial attribution annotations. 

Ego4D-M has annotations for semantic, spatial, and temporal attribution where the PNR frame numbers and the mistake grounding box are inherited from the original Ego4D dataset. 
For samples with only predicate mistakes, we use the union of the hand regions as the spatial attribution annotation. For samples with only object mistakes, we use the original annotated object bounding boxes as the spatial attribution annotation. For samples with both predicate and object mistakes, we take the union of the hand and object regions as the spatial attribution annotation.
EPIC-KITCHENS-M only has semantic annotation due to the lack of PNR frame number annotation in the original dataset.

\cref{tab:dataset-comparison} contains a comparison of existing mistake datasets, both used egocentric action datasets, and the new MisEngine-produced variants, Ego4D-M and EPIC-KITCHENS-M.  The produced MATT datasets are at least two orders of magnitude larger than any of the existing mistake datasets in the literature.

\begin{table}[t]
    \centering
    \resizebox{\columnwidth}{!}{%
        \begin{tabular}{lccccccccccc}
        \toprule
        & \multicolumn{2}{c}{\textbf{\# Samples}}
        & \multicolumn{2}{c}{\textbf{Semantic Var.}}
        & \multicolumn{2}{c}{\textbf{Visual Var.}}
        & \multicolumn{4}{c}{\textbf{Annotation}} \\
        \cmidrule(lr){2-3} \cmidrule(lr){4-5} \cmidrule(lr){6-7} \cmidrule(lr){8-11}
        \textbf{Dataset}
        & \textbf{Total} & \textbf{By activity}
        & \textbf{Activ.} & \textbf{Dm.}
        & \textbf{Env.} & \textbf{Part.}
        & \textbf{Det.} & \textbf{Sem.} & \textbf{Temp.} & \textbf{Spa.} \\
        \midrule
        EgoPER~\cite{lee2024error} & 599 & 9.7 & 62 & 1 & 2 & 11 &  \cmark & \pmark & \xmark & \pmark\\
        Assembly101~\cite{sener2022assembly101, ding2023every} & 707 & 2.0 & 358 & 1 & 1 & 53 & \cmark & \pmark & \xmark & \xmark \\
        HoloAssist~\cite{wang2023holoassistegocentrichumaninteraction} & 7,562 & 0.91 & 8,285 & 2 &  - & 222 & \cmark & \pmark & \pmark & \xmark \\
        CaptionCook4D~\cite{peddi2024captaincook4d} & 1,964 & 5.6 & 352 & 1 & 10 & 8 & \cmark & \pmark & \xmark & \xmark  \\
        Epic-Tent~\cite{jang2019epic} & 626 & 52.2 & 12 & 1 & - & 24 & \cmark & \pmark & \xmark & \pmark \\
        \midrule
        EPIC-KITCHENS & 89,975 & 5.0 & 18,003 & 1 & 45 & 37 & \xmark  & \xmark & \xmark & \pmark \\
        Ego4D & 155,367 & 2.1 & 75,423 & 14+ & 100+  & 931 & \xmark  & \xmark & \pmark & \pmark \\
        \midrule
        EPIC-KITCHENS-M & 221,094 & 18.0 & 12,283 & 1 & 45 & 37 & \cmark  & \cmark & \xmark & \pmark \\
        Ego4D-M & 257,584 & 16.0 & 16,099 & 14+ & 100+  & 248 & \cmark  & \cmark & \cmark & \cmark \\
        \bottomrule
        \end{tabular}
    }
    \caption{Statistics of egocentric video normal and mistake datasets. 
    Our 
    \# Samples are the small video segments that contain the mistake instead of the video of the whole procedure. \# Activ. are the number of unique short actions in the dataset. Dm. is the number of domains in the dataset (e.g., cooking, cleaning, etc.). Env. is the number of unique environments when collecting in the whole dataset. \# Part. are the number of unique individuals that perform the activities in the dataset. Annotation presence columns: (Det)ection, (Sem)antic, (Temp)oral PNR, (Spa)tial bounding box.  The \pmark~denotes the annotation is partially available (e.g., natural language as semantic explanation for mistakes).
    }
    \vspace{-1em}
    \label{tab:dataset-comparison}
\end{table}

\begin{figure*}
    \centering
    \includegraphics[width=1\textwidth]{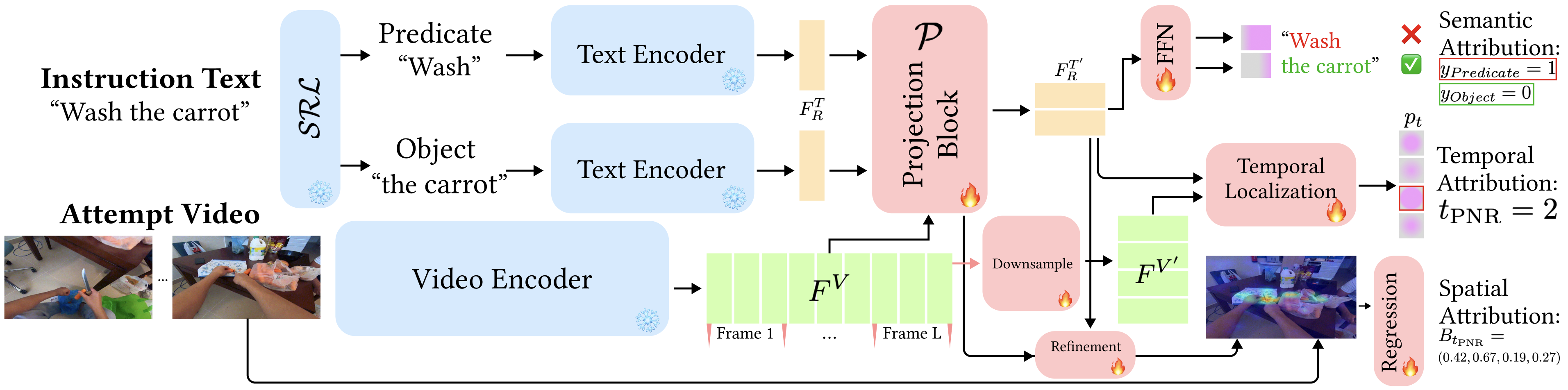}
    \caption{
    MisFormer’s unified architecture for mistake attribution. MisFormer jointly processes the instruction text and an attempt video, extracting shared multimodal features. Three specialized transformer heads perform semantic attribution (detecting misaligned  roles), temporal localization (pinpointing the Point-of-No-Return frame), and spatial localization (predicting mistake regions via attention-driven bounding boxes), enabling comprehensive and interpretable mistake analysis across semantic, temporal, and spatial dimensions.}
    \label{fig:architecture_model}
\end{figure*}
\subsection{MisFormer: A unified model for fine-grained mistake understanding}
\label{sec:model}

We propose a unified model structure, MisFormer ($\mathcal{F}$), to jointly address all three mistake attribution tasks: semantic, temporal, and spatial localization. 
As illustrated in Figure~\ref{fig:architecture_model}, given an instruction text $T$ and an attempt video $V \in \mathds{R}^{L \times H \times W \times 3}$, MisFormer first extracts multimodal features through a shared feature extraction module, then processes them through three specialized attribution heads to produce the outputs for each subtask: semantic attribution labels $(y_r)_{r \in \mathcal{R}}$, Point-of-No-Return frame $t_{\mathrm{PNR}}$, and mistake bounding box $B_{t_{\mathrm{PNR}}}$.

\paragraph{Feature Extraction}
We begin by processing the instruction text $T$ using AllenNLP's Semantic Role Labeling~\cite{allen_srl} to decompose it into semantic role substrings $S_r$ for each role $r \in \mathcal{R}$.
For each $S_r$, we extract text features $F_r^{T}$ using the text encoder from InternVideo2~\cite{internvideo2}, yielding $F_R^{T} = \{F_r^{T}\}_{r \in \mathcal{R}} \in \mathds{R}^{|\mathcal{R}| \times d}$ where each semantic role is represented as a $d$-dimensional embedding. 
In parallel, we extract video features $F^{V} \in \mathds{R}^{L \times K \times d}$ from the attempt video using InternVideo2's video encoder, where $K$ is the number of spatial patches per frame. 

Due to InternVideo2's video-language pre-training, $F_r^{T}$ and $F^{V}$ naturally reside in a shared embedding space where text features are semantically aligned with corresponding visual patches.
To further adapt these features for mistake understanding, we employ a projection block $\mathcal{P}$ consisting of 2 Transformer decoder layers without causal masking for self-attention. 
Within $\mathcal{P}$, self-attention operates over the text features $F_R^{T}$ (treating each $F_r^{T}$ as a single token), enabling information exchange across semantic roles. 
Cross-attention then attends from $F_R^{T}$ (as queries) to $F^{V}$ (as keys and values), connecting visual context to each semantic role. 
The output is projected text features $F_R^{T'} \in \mathds{R}^{|\mathcal{R}| \times d}$, which encodes both inter-role relationships and video-grounded semantics.

\paragraph{Semantic Attribution}
The semantic attribution head determines which semantic roles in the instruction are executed incorrectly in the attempt. 
For each role $r \in \mathcal{R}$, we apply a feed-forward network (FFN)~\cite{DBLP:journals/corr/VaswaniSPUJGKP17} followed by a sigmoid activation to the projected text features $F_r^{T'}$, producing a binary prediction $\hat{y}_r \in \{0, 1\}$ indicating whether the role is misaligned. 
We train this module with a binary cross-entropy loss $\mathcal{L}_S$ aggregated across all semantic roles:
$
\mathcal{L}_S = - \sum_{r \in \mathcal{R}} \left[ y_r \log(\hat{y}_r) + (1-y_r) \log(1-\hat{y}_r) \right],
$
where $y_r$ is the ground-truth label for role $r$. 

\paragraph{Temporal Attribution}
The temporal attribution head localizes the Point-of-No-Return (PNR) frame where the mistake becomes irreversible. 
We first apply a downsampling block consisting of 2 self-attention layers to aggregate spatial tokens within each frame of $F^{V}$, producing frame-level features $F^{V'} \in \mathds{R}^{L \times d}$. 
These frame features are then processed by a temporal localization block with 2 Transformer decoder layers (without causal masking), where cross-attention operates with $F^{V'}$ as queries and $F_R^{T'}$ as keys and values. 
An FFN with output dimension 1 followed by a softmax over the temporal dimension produces a probability distribution $p_t \in [0, 1]$ for each frame $t \in [L]$. 
The PNR frame is identified as $t_{\mathrm{PNR}} = \arg\max_{t} p_t$.

We supervise this module with a cross-entropy loss:
$
\mathcal{L}_T = - \sum_{t=1}^{L} y_t \log(p_t),
$
where $y_t \in \{0, 1\}$ is the ground-truth label indicating the PNR frame. 
During training, $\mathcal{L}_T$ is computed only for samples annotated as mistakes. 
During inference, temporal localization is performed only for samples where semantic attribution identifies at least one misaligned role.

\paragraph{Spatial Attribution}
The spatial attribution head identifies the mistake region within the PNR frame by predicting a bounding box $B_{t_{\mathrm{PNR}}}$. 
To localize the relevant spatial regions, we extract the cross-attention weights $A_{r, t_{\mathrm{PNR}}} \in \mathds{R}^{K \times d}$ from the final cross-attention layer of the projection block~$\mathcal{P}$, where $K$ is the number of spatial patches. 
We apply refinement block to, first concatenate $F_r^{T'}$ to $A_{r, t_{\mathrm{PNR}}}$ in feature dimension of all tokens, then apply two self-attention blocks to produce spatial attention scores, which are reshaped to a 2D spatial map and upsampled via bilinear interpolation to match the frame resolution, forming a visual saliency heatmap.

This heatmap is then concatenated with the RGB PNR frame as a 4-channel input to a lightweight CNN-based regression module that predicts the bounding box coordinates. 
When multiple semantic roles are identified as misaligned, we merge their attention maps and produce a unified bounding box encompassing all relevant mistake regions.
We supervise the spatial attribution head with Huber loss~\cite{huber1964robust} between the predicted and ground-truth bounding boxes.

\section{Experiments}
\label{sec:experiments}
\subsection{Setup}
\paragraph{Baselines}
To the best of our knowledge, no prior work covers all of MATT's specifics. We therefore adapt baselines separately for semantic, temporal, and spatial attribution, as well as for mistake detection.
For semantic attribution, we evaluate ChatGPT-4o~\cite{openai2024gpt4o}, fine-tuned Video-ChatGPT~\cite{maaz-etal-2024-video}, and LLaVA-NeXT Video~\cite{zhang2024llavanextvideo} by prompting them to decide ``mistake'' or ``correct'' for each semantic role.
For ChatGPT-4o, we evaluate on 1000 random samples from the test sets.
For temporal attribution, we compare with fine-tuned SOTA Point-of-No-Return (PNR) temporal localization methods, EgoMotion‑COMPASS~\cite{lei2022masked} and EgoT2~\cite{xue2023egot2} (with PNR as target task). 
For spatial attribution, we compare with MediaPipe~\cite{mediapipe} and fine-tuned SSDA~\cite{leonardi2024synthetic}, SOTA hand/hand–object interaction detectors, to segment the hands and object in the PNR frame.
We also compare against SOTA mistake detection methods, AMNAR~\cite{AMNAR_Huang_2025_CVPR} and EgoPED~\cite{lee2024error}. 
Our mistake detection prediction is derived by aggregating semantic attribution: a clip is labeled as a mistake if any semantic role is flagged as such. 

\paragraph{Evaluation Metrics}
For semantic attribution, we treat each semantic role as a binary classification problem. Following prior work~\cite{lee2024error}, we report F1@0.5 and Accuracy as the main metrics, both per semantic role and averaged across roles.
Following the PNR localization task~\cite{grauman2022ego4d}, we evaluate temporal attribution by Mean Absolute Error, both per frame and per second.
For spatial attribution, we report mean Intersection-over-Union (mIoU) between the predicted and ground-truth boxes, and additionally report Center Distance (CD) and Box Size Error (BSE).
For mistake detection, we follow common practice~\cite{xue2023egot2} and report F1@0.5 and accuracy.

\paragraph{Dataset}
We benchmark attribution subtasks on Ego4D-M and EPIC-KITCHENS-M. We split each dataset into training/validation/test sets in an 8:1:1 ratio. This yields 206K/25K/25K for Ego4D-M and 176K/22K/22K samples for EPIC-KITCHENS-M. 
We additionally use EgoPER~\cite{lee2024error}, an existing mistake dataset, for mistake detection evaluation.

\subsection{Results}
\label{sec:main_results}
\paragraph{Mistake Semantic Attribution}
For both datasets, on average (see the ``Average" column in~\cref{tab:sem_att_results}), fine-tuned open-source video-language models (LLaVA-Vid~\cite{zhang2024llavanextvideo} and Vid-Chat~\cite{maaz-etal-2024-video}) perform poorly, staying below 75\% F1 on EPIC-KITCHENS-M and below 45\% F1 on Ego4D-M. 
ChatGPT, a commercial closed-source model likely trained on larger corpora, reaches 77.23\% and 50.95\% F1 on EPIC-KITCHENS-M and Ego4D-M, respectively. 
MisFormer attains 83.89\% and 56.24\% F1 on EPIC-KITCHENS-M and Ego4D-M, surpassing the best baseline by 6.66\% and 5.30\%, respectively. 

Role-wise, MisFormer improves Predicate by 10.60\% and 5.02\% F1 on EPIC-KITCHENS-M and Ego4D-M, respectively. For Object, the gains are 1.99\% and 5.55\% F1 on the two datasets. 
Across methods, Object consistently outperforms Predicate (verbs), highlighting the challenge of modeling fine-grained motions in egocentric videos, consistent with prior findings~\cite{egovlp2022}.
Accuracy mirrors the F1 score trends, as our datasets are approximately class-balanced. 

\begin{table}[h!]
    \centering
    \resizebox{\columnwidth}{!}{%
    \begin{tabular}{cccccccc}
        \toprule
        \multirow{2}{*}{\textbf{Dataset}} & \multirow{2}{*}{\textbf{Method}} & \multicolumn{2}{c}{\textbf{Average}} & \multicolumn{2}{c}{\textbf{Predicate}} & \multicolumn{2}{c}{\textbf{Object}} \\
        \cmidrule(r){3-4} \cmidrule(r){5-6} \cmidrule(r){7-8}
        & & \textbf{Acc. $\uparrow$} & \textbf{F1 $\uparrow$} & \textbf{Acc. $\uparrow$} & \textbf{F1 $\uparrow$} & \textbf{Acc. $\uparrow$} & \textbf{F1 $\uparrow$} \\
        \multirow{4}{*}{\textbf{EK}} 
        & LLaVA-Vid~\cite{zhang2024llavanextvideo} & 71.71 & 70.34 & 64.10 & 64.55 & 79.32 & 76.13 \\
        & Vid-Chat~\cite{maaz-etal-2024-video} & 64.17 & 63.27 & 54.45 & 52.11 & 73.89 & 74.43 \\
        & ChatGPT~\cite{openai2024gpt4o} & 77.66 & 77.23 & 66.23 & 65.11 & 89.09 & 89.35 \\
        & \cellcolor{gray!20} MisFormer (Ours)$^\dagger$ & \cellcolor{gray!20} \textbf{84.91} & \cellcolor{gray!20} \textbf{84.78} & \cellcolor{gray!20} \textbf{76.35} & \cellcolor{gray!20} \textbf{75.31} & \cellcolor{gray!20} \textbf{93.47} & \cellcolor{gray!20} \textbf{94.25} \\
        & \cellcolor{gray!20} MisFormer (Ours) & \cellcolor{gray!20} \textbf{84.13} & \cellcolor{gray!20} \textbf{83.89} & \cellcolor{gray!20} \textbf{76.83} & \cellcolor{gray!20} \textbf{76.43} & \cellcolor{gray!20} \textbf{91.43} & \cellcolor{gray!20} \textbf{91.34} \\
    \midrule
    \multirow{4}{*}{\textbf{Ego}}
        & LLaVA-Vid~\cite{zhang2024llavanextvideo}  & 42.83 & 40.12 & 48.10 & 47.95 & 37.56 & 32.29 \\
        & Vid-Chat~\cite{maaz-etal-2024-video} & 36.47 & 34.36 & 36.73 & 36.84 & 36.21 & 31.88 \\
        & ChatGPT~\cite{openai2024gpt4o} & 52.45 & 50.95 & 53.42 & 50.20 & 51.48 & 51.70 \\
        & \cellcolor{gray!20} MisFormer (Ours)$^\dagger$ & \cellcolor{gray!20} \textbf{59.37} & \cellcolor{gray!20} \textbf{55.41} & \cellcolor{gray!20} \textbf{55.44} & \cellcolor{gray!20} \textbf{53.03} & \cellcolor{gray!20} \textbf{63.33} & \cellcolor{gray!20} \textbf{57.79} \\
        & \cellcolor{gray!20} MisFormer (Ours) & \cellcolor{gray!20} \textbf{62.03} & \cellcolor{gray!20} \textbf{56.24} & \cellcolor{gray!20} \textbf{58.40} & \cellcolor{gray!20} \textbf{55.22} & \cellcolor{gray!20} \textbf{65.66} & \cellcolor{gray!20} \textbf{57.25} \\
\bottomrule
    \end{tabular}
    }
    \caption{Semantic attribution results. MisFormer consistently outperforms the SOTA Video-Language model baselines across all datasets on all semantic roles. $\dagger$ denotes the model is evaluated on the subset of test set used by ChatGPT baseline for fair comparison.}
    \label{tab:sem_att_results}
\end{table}

\vspace{-10pt}
\paragraph{Mistake Temporal Attribution}
Quantitative results on Ego4D-M are shown in~\cref{tab:temporal_attribution}. Compared to the best existing method, MisFormer reduces mean absolute error (MAE) by 21.81\%---a decrease of 5.34 frames and 0.178 seconds. 
For reference, the average clip length is about 8 seconds (240 frames at 30 fps), so this corresponds to a 2.23\% reduction relative to video duration. 
Whereas prior PNR localization methods rely solely on video signals, MisFormer additionally conditions on instruction text. Even when text-video alignment is imperfect, this supervision provides complementary cues that improve temporal localization. 

\begin{table}[h!]
    \centering
    \resizebox{\columnwidth}{!}{%
    \begin{tabular}{ccc}
        \toprule
            \textbf{Method} & \textbf{MAE (frames) $\downarrow$} & \textbf{MAE (seconds) $\downarrow$} \\
            \midrule
            EgoMotion-COMPASS~\cite{lei2022masked} & 48.96 & 1.632 \\
            EgoT2~\cite{xue2023egot2} & 24.48 & 0.816 \\
            \rowcolor{gray!20} MisFormer (Ours) & \textbf{19.14} & \textbf{0.638} \\
            \bottomrule
        \end{tabular}
    }
    \caption{Temporal attribution on Ego4D-M. MisFormer reduces mean absolute error (MAE) by at least 21.81\% compared to the SOTA PNR localization methods, which corresponds to a 2.23\% reduction relative to the attempt video duration.}
    \label{tab:temporal_attribution}
\end{table}

\begin{figure}
    \centering
    \resizebox{\columnwidth}{!}{%
    \includegraphics[width=1\textwidth]{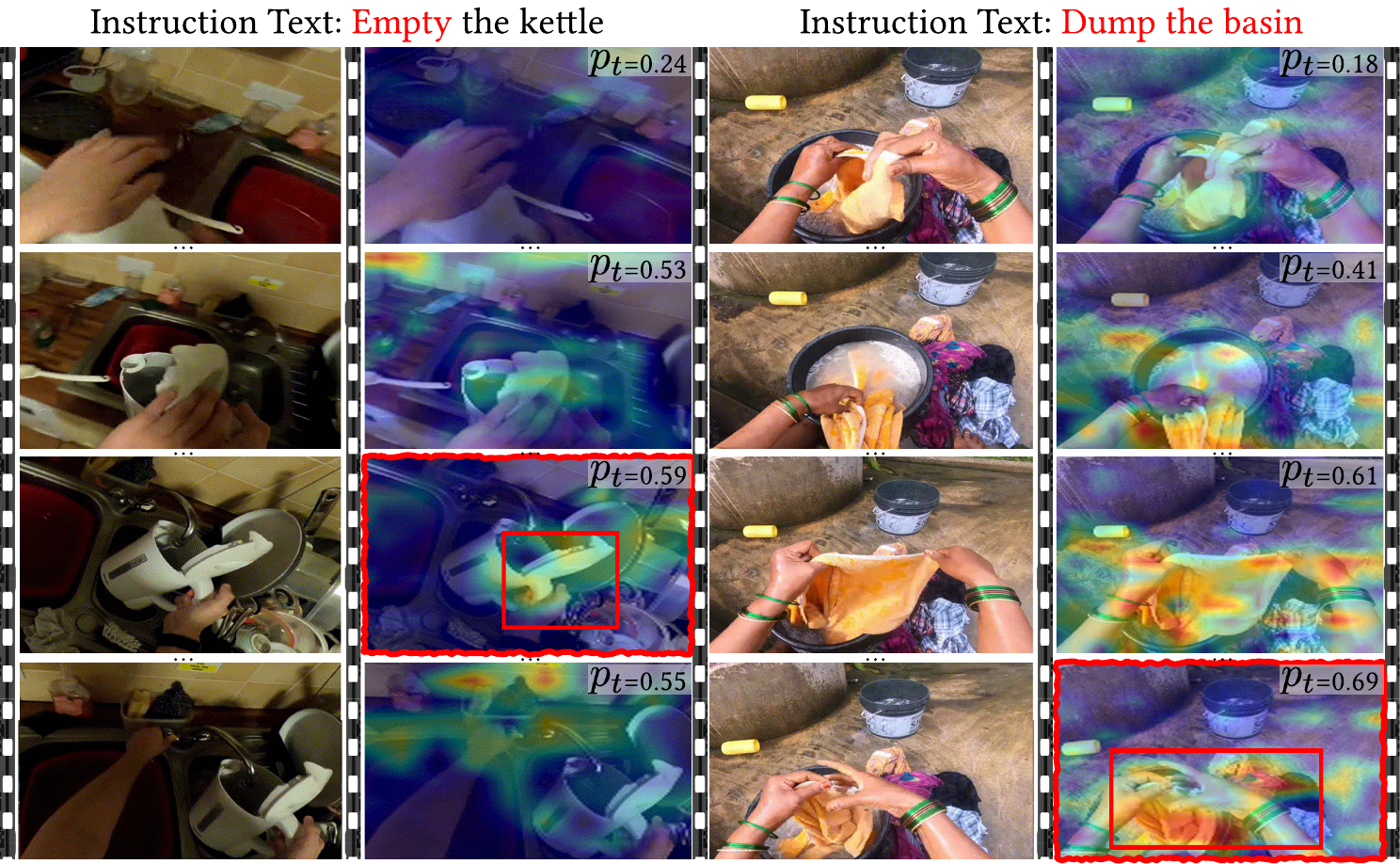}
    }
    \caption{
        Qualitative results of MisFormer. Red text highlights mistaken semantic roles, the red frame marks the Point-of-No-Return (PNR), and the red bounding box localizes the mistake region in the PNR frame. The column in each sample visualizes the per-frame heatmap from the spatial attribution module for reference.
        } 
    \label{fig:grounding}
\end{figure}
\vspace{-10pt}
\paragraph{Mistake Spatial Attribution}
Hand-detection methods fail to address the spatial attribution task because they do not localize the manipulated object, which is important for precise mistake grounding and downstream recovery~\cite{levine2018learning}.
MisFormer substantially outperforms hand-detection baselines (MediaPipe-U) for spatial attribution on the PNR frame, improving mIoU by +9.33 and reducing Center Distance~(CD) and Box Size Error~(BSE) by 3.11 and 4.71, respectively. 
Compared to a SOTA hand-object interaction detector (SSDA), MisFormer is less accurate on localization (mIoU 59.21 vs. 68.54; CD 10.36 vs. 4.52; BSE 16.27 vs. 8.56). Nevertheless, MisFormer offers a unified framework which performs semantic and temporal attribution in conjunction with localization. 
MisFormer is also efficient. Spatial attribution module runs at 68.9 FPS versus 28.44 FPS (SSDA) on the same NVIDIA H100 GPU.

\begin{table}[h!]
    \centering
    \resizebox{\columnwidth}{!}{%
        \begin{tabular}{cccc}
            \toprule
            \textbf{Method} & \textbf{mIoU (\%) $\uparrow$} & \textbf{CD (\%) $\downarrow$} & \textbf{BSE (\%) $\downarrow$} \\
            \midrule
            MediaPipe-U~\cite{mediapipe} & 49.88 & 13.47 & 20.98 \\
            SSDA~\cite{leonardi2024synthetic} & \textbf{64.54} & \textbf{7.21} & \textbf{12.34}  \\
            \rowcolor{gray!20} MisFormer (Ours) & \underline{59.21} & \underline{10.36} & \underline{16.27} \\
            \bottomrule
        \end{tabular}
    }
    \caption{Spatial attribution results. We report mIoU, Center Distance (CD), and Box Size Error (BSE). MisFormer, our unified framework, performs better than the hand-detection baselines (by 18.70\% mIoU) and comparable to hand-object interaction detector. \textbf{Bold} and \underline{underline} indicate the best and second best results.}
    \label{tab:spatial_attribution}
\end{table} 

\paragraph{Mistake Detection}
Semantic attribution implies mistake detection, so we also compare MisFormer with state-of-the-art mistake detection methods on our MATT benchmarks (EPIC-KITCHENS-M, Ego4D-M) and on the existing popular Mistake Detection benchmark EgoPER~\cite{lee2024error}. 
In \cref{tab:mistake_detection}-(a), prior methods perform poorly on our benchmarks because they are designed for a small set of activities with no mistake supervision, casting detection as out-of-distribution (OOD) detection problem. 
When there are too many activities todetect mistakes on, a single OOD model struggles to separate actual mistakes from benign variability (i.e., correct samples in many activities). 
In contrast, MisFormer scales with data via supervised learning, consistent with findings on data-driven approaches in other domains~\cite{kaplan2020scalinglawsneurallanguage,sun2017revisiting}.

However, such models struggle when there is lack of training data and supervision signal. In \cref{tab:mistake_detection}-(b), training MisFormer from scratch on an augmented EgoPER training set using only semantic attribution supervision underperforms specialized detectors. 
Notably, such struggles can be alleviated by pretraining. Row \textit{MisFormer (Ours)$^\dagger$} retrains MisFormer on EPIC-KITCHENS-M and then fine-tunes on EgoPER, yielding competitive results with SOTA detectors while preserving our unified attribution capabilities. 
These results suggest that MisFormer can handle mistake understanding ranging from coarse detection to fine-grained attribution when fueled by the large data prepared by MisEngine. 

\begin{table}[h!]
    \centering
    \begin{minipage}{0.59\columnwidth}
        \centering
        \resizebox{\linewidth}{!}{%
        \begin{tabular}{ccccc}
            \toprule
            \multirow{2}{*}{\textbf{Method}} & \multicolumn{2}{c}{\textbf{EK}} & \multicolumn{2}{c}{\textbf{Ego}} \\
            \cmidrule(lr){2-3}\cmidrule(lr){4-5}
            & \textbf{F1@.5 $\uparrow$} & \textbf{Acc. $\uparrow$} & \textbf{F1@.5 $\uparrow$} & \textbf{Acc. $\uparrow$} \\
            \midrule
            AMNAR~\cite{AMNAR_Huang_2025_CVPR} & 16.34 & 17.45 & 12.57 & 14.24 \\
            EgoPED~\cite{lee2024error} & 21.18 & 22.23 & 15.62 & 17.34 \\
            \rowcolor{gray!20} MisFormer (Ours) & \textbf{78.05} & \textbf{80.90} & \textbf{57.55} & \textbf{59.89} \\
            \bottomrule
        \end{tabular}%
        }
        \vspace{3pt}
        \small \textbf{(a)} 
    \end{minipage}
    \hfill
    \begin{minipage}{0.39\columnwidth}
        \centering
        \resizebox{\linewidth}{!}{%
        \begin{tabular}{ccc}
            \toprule
            \textbf{Method} & \textbf{F1@.5 $\uparrow$} & \textbf{Acc. $\uparrow$} \\
            \midrule
            AMNAR~\cite{AMNAR_Huang_2025_CVPR} & 28.54 & 29.65 \\
            EgoPED~\cite{lee2024error} & 32.18 & 33.23 \\
            \rowcolor{gray!20} MisFormer (Ours) & 26.78 & 27.89 \\
            \rowcolor{gray!20} MisFormer (Ours)$^\dagger$ & \textbf{35.18} & \textbf{36.23} \\
            \bottomrule
        \end{tabular}%
        }
        \vspace{3pt}
        \small \textbf{(b)} 
    \end{minipage}
    \caption{Mistake Detection results on: (a) EPIC-KITCHENS-M (EK) and Ego4D-M (Ego); (b) EgoPER dataset. When fueled by large data prepared by MisEngine, MisFormer can handle mistake understanding ranging from coarse detection to fine-grained attribution. $\dagger$ denotes the model is pretrained on EPIC-KITCHENS-M. }
    \label{tab:mistake_detection}
\end{table}
\vspace{-10pt}
\paragraph{Ablation Study}
In \cref{tab:ablation}, we ablate key design choices of MisFormer across the feature extractor (\textbf{b}), semantic attribution (\textbf{c}), temporal attribution (\textbf{d}), and spatial attribution (\textbf{e}) modules. 
In \textbf{(b)}, we replace the backbone with another video-language pre-trained model, LaViLa~\cite{Zhao_2023_CVPR}, as the feature extractor. 
Performance drops across all tasks when using LaViLa. In contrast, InternVideo2~\cite{internvideo2} (our default) is pre-trained with more modalities, broader objectives, and more data, highlighting the importance of high-capacity video-language pre-training for MATT. 
In \textbf{(c)}, we remove the projection block $\mathcal{P}$ and feed the pre-trained text encoder features $F_R^{T}$ directly into the attribution heads. 
Without this projection step, raw text embeddings are insufficient to capture the nuanced divergence between the instruction and the attempt video, degrading performance. 
In \textbf{(d)}, we remove supervision for temporal attribution. At inference, we run spatial attribution on all frames and select the frame with the highest average activation in the score map as the PNR prediction. 
This has little effect on semantic attribution and mistake detection, but clearly degrades temporal attribution quality, demonstrating the benefit of using temporal attribution as a training objective. 
In \textbf{(e)}, we generate temporal heatmaps with GradCAM~\cite{gradcam2017} and otherwise follow the same spatial decoding procedure. 
The inferior spatial results confirm the benefit of using a refined attention-based heatmap for retrieving the most relevant mistake regions. 

\begin{table}[h!]
    \centering
    \resizebox{\columnwidth}{!}{%
        \begin{tabular}{cccccc}
            \toprule
            \textbf{No.} & \textbf{Method} & \textbf{Sema. F1@0.5 $\uparrow$} & \textbf{Temp. MAE (s) $\downarrow$} & \textbf{Spat. mIoU (\%) $\uparrow$} & \textbf{Det. F1@0.5 $\uparrow$} \\
            \midrule
            \rowcolor{gray!20} \textbf{(a)} & MisFormer (Ours) & \textbf{56.24} & \textbf{0.438} & \textbf{59.21} & \textbf{57.55}  \\
            \textbf{(b)} & Ours-LaViLa~\cite{Zhao_2023_CVPR} & 49.16 & 0.561 & 51.37 & 46.05 \\
            \textbf{(c)} & Ours-w/o $\mathcal{P}$ & 51.34 & 0.457 & 55.43 & 52.75 \\
            \textbf{(d)} & Ours-w/o Temp. & 51.29 & 0.623 & 57.78 & 57.46 \\
            \textbf{(e)} & Ours-GradCAM~\cite{gradcam2017} & 55.52 & 0.482 & 55.03 & 57.51 \\
            \bottomrule
        \end{tabular}
    }
    \caption{Ablations on Ego4D‑M for Semantic Attribution (Sema.), Temporal Attribution (Temp.), Spatial Attribution (Spat.), and Mistake Detection (Det.). Our default model is MisFormer (Ours). Results show the importance of the design choices in each module.}
    \label{tab:ablation}
\end{table}
\vspace{-10pt}
\section{Discussion}
\label{sec:conclusion}

\paragraph{Conclusion}
We introduced Mistake Attribution (MATT), focusing on what, when, and where mistakes arise in egocentric videos. 
We created large-scale, attribution-rich datasets with MisEngine (Ego4D-M, EPIC-KITCHENS-M) by leveraging existing egocentric action corpora.
Fueled by these datasets, we developed MisFormer, a unified model that attributes mistakes along the semantic, temporal, and spatial axes.
Experiments show that existing methods---even when specialized for individual subtasks---struggle to address MATT, underscoring its challenge.
MisFormer consistently outperforms strong video-language, temporal localization, hand-object interaction and mistake detection baselines on our and prior benchmarks, demonstrating the effectiveness of our data-driven solution for MATT.

\paragraph{Limitations and Future Work}
Real applications will involve longer, compositional instructions with richer role inventories. 
As our experiments indicate, different roles pose distinct challenges; study the challenges and solutions for MATT with longer instructions is an important next step. 
Second, MisFormer currently relies on features from existing video–language representation learning, which is pre-trained with video-language alignment objective. While our downstream adaptation blocks make these features effective for MATT, developing a task-driven representation that jointly embeds video and instruction specifically for mistake understanding could further improve attribution accuracy and efficiency, potentially serve as a foundation for mistake understanding in general.

\section*{Acknowledgement}
This research was funded, in part, by the U.S. Government under ARPA-H contract 1AY2AX000062. The views and conclusions contained in this document are those of the authors and should not be interpreted as representing the official policies, either expressed or implied, of the U.S. Government.

{
    \small
    \bibliographystyle{ieeenat_fullname}
    \bibliography{main}
}

\clearpage

\twocolumn[{
    \begin{center}
        \textbf{\Large Mistake Attribution: Fine-Grained Mistake Understanding in Egocentric Videos}\\[0.4em]
        \large Supplementary Material
    \end{center}
    \vspace{1em}
}]

\setcounter{section}{0}
\renewcommand{\thesection}{\Alph{section}}

\section{MisEngine Details}
\vspace{-8pt}
\noindent \textbf{Role-matching function $\mathcal{J}$.}
$\mathcal{J}(g^r_i, g^r_j)$ determines whether role $r$ matches between two action descriptions, and thus whether a cross-matched video contains a mistake on that role. The matching rule is granularity-dependent: e.g., ``cut'' vs.\ ``dice'' are distinct in a professional kitchen but interchangeable in casual cooking. We implement two variants(\cref{sec:dataset-construction}): (1)~\emph{taxonomy-based} $\mathcal{J}$ for Ego4D, grouping synonymous verbs and nouns from the original dataset; and (2)~\emph{literal string matching} for EPIC-KITCHENS. SRL on 1K Ego4D descriptions yields 97.97\% predicate and 94.13\% object F1, reflecting the short imperative phrasing of the source annotations.

\vspace{-10pt}
\section{Dataset Ecological Validation}
\vspace{-8pt}
We conduct a user study ($16$ participants, $200$ $\langle$text, video$\rangle$ pairs per dataset, no prior exposure) on 4 datasets to evaluate how realistic our constructed mistake samples and inherited annotations are. We report Real-\%: the \% of participants who judge the sample pair or inherited annotation as realistic, averaged over videos.

\vspace{-2mm}
\smallskip
\begin{table}[h]
\centering
\vspace{-2pt}
\scriptsize
\setlength{\tabcolsep}{2pt}
\begin{subtable}{\linewidth}
\centering
\resizebox{\linewidth}{!}{%
\begin{tabular}{cccc}
\toprule
EK-M (ours)  & Ego4D-M (ours) & EgoPER~[\underline{23}] & CC4D~[\underline{37}] \\
\midrule
92.32$\pm$6.04 & 92.42$\pm$7.01 & 94.17$\pm$4.02 & 95.75$\pm$4.03 \\
\bottomrule
\end{tabular}}
\vspace{-1mm}
\caption{\% of samples human-rated as realistic (Real-\% $\uparrow$) on mistake videos.}
\label{tab:realism}
\end{subtable}
\\[1mm]

\begin{subtable}[t]{0.50\linewidth}
\centering
\resizebox{\linewidth}{!}{%
\setlength{\tabcolsep}{2pt}
\begin{tabular}{@{}cc@{}}
\toprule
Tmp.\ (\%) & Spa.\ (\%) \\
\midrule
96.29 $\pm$ 3.99 & 95.81 $\pm$ 2.16 \\
\bottomrule
\end{tabular}}
\vspace{-1mm}
\caption{Real-\% on inherited annotations.}
\label{tab:anno_quality}
\end{subtable}
\begin{subtable}[t]{0.49\linewidth}
\centering
\resizebox{\linewidth}{!}{%
\setlength{\tabcolsep}{2pt}
\begin{tabular}{@{}ccc@{}}
\toprule
$\text{Mis}_{V}$ & $\text{Mis}_{Obj}$ & $\text{Mis}_{Both}$ \\
\midrule
60.73 & 57.99 & 58.91 \\
\bottomrule
\end{tabular}}
\vspace{-1mm}
\caption{{\fontsize{6}{8}\selectfont Spatial-Attribution mIoU by mistake type.}}
\label{tab:iou_by_type}
\end{subtable}%

\vspace{-2mm}
\end{table}

\vspace{-1mm}
\noindent
Tab.~(a): All four datasets receive comparably high realism ratings (within std). This is expected since cross-matched samples are drawn from the same domain, grounded in original metadata. Participants occasionally flag samples in prior datasets as ``staged,'' consistent with the manual collection challenges discussed in \cref{sec:intro}.

\vspace{-1mm}
\smallskip \noindent
Tab.~(b): Participants judge whether the inherited PNR frame marks the consolidation of a mistake (Tmp.) and whether the bounding box captures mistake-relevant detail (Spa.). Both achieve $>$95\% Real-\%, validating the quality of inherited annotations. The temporal attribution frame occurs at $79.48$\% of video duration on average. Samples are constructed in equal numbers across mistake types for balanced training; this count is configurable (\cref{sec:dataset-construction}).
\vspace{-10pt}
\section{Additional Experiments}
\vspace{-8pt}
\noindent \textbf{Zero-shot transfer to EgoPER.}
\MODEL{} generalizes zero-shot to EgoPER, achieving $34.46$ F1 / $35.52$ Acc and surpassing both baselines.~\cref{tab:mistake_detection} results show that manually collected datasets provide insufficient training signal for \MODEL{} (\cref{tab:dataset-comparison}), underscoring the necessity of \DATAP{}.

\vspace{-1mm}
\smallskip
\noindent \textbf{Spatial attribution mIoU by mistake type.}
Tab.~(c) reports spatial attribution mIoU broken down by mistake type. $\text{Mis}_\text{Both}$ (which uses a union bounding box over hand and object) performs comparably to single-role types. Union boxes remain compact because, at the PNR frame, hand and object regions substantially overlap during active interaction.

\vspace{-1mm}
\smallskip
\noindent \textbf{Ablation: semantic head and projection block.}~\cref{tab:ablation} in the main paper ablates the projection block $\mathcal{P}$ and task-specific heads. Removing $\mathcal{P}$ degrades all tasks (row c). Removing temporal or spatial supervision each hurts the complementary task (rows d, e), confirming their mutual benefit. We additionally ablate the semantic head: without it, semantic attribution and mistake detection are disabled, while temporal and spatial attribution improve to $0.358$\,s MAE and $65.33$\% mIoU. This suggests the current $\mathcal{P}$ faces tension between representing semantic-level and grounded-level features simultaneously, which we identify as a direction for future work. The unified structure nonetheless avoids training separate models, runs $2.42\times$ faster (\cref{sec:main_results}), and outperforms 6 of 7 baselines (\cref{tab:sem_att_results,tab:spatial_attribution,tab:temporal_attribution}).

\vspace{-10pt}
\section{Qualitative Results and Failure Analysis}
\begin{figure}[h]
    \centering
    \vspace{-9pt}
    \includegraphics[width=1\linewidth]{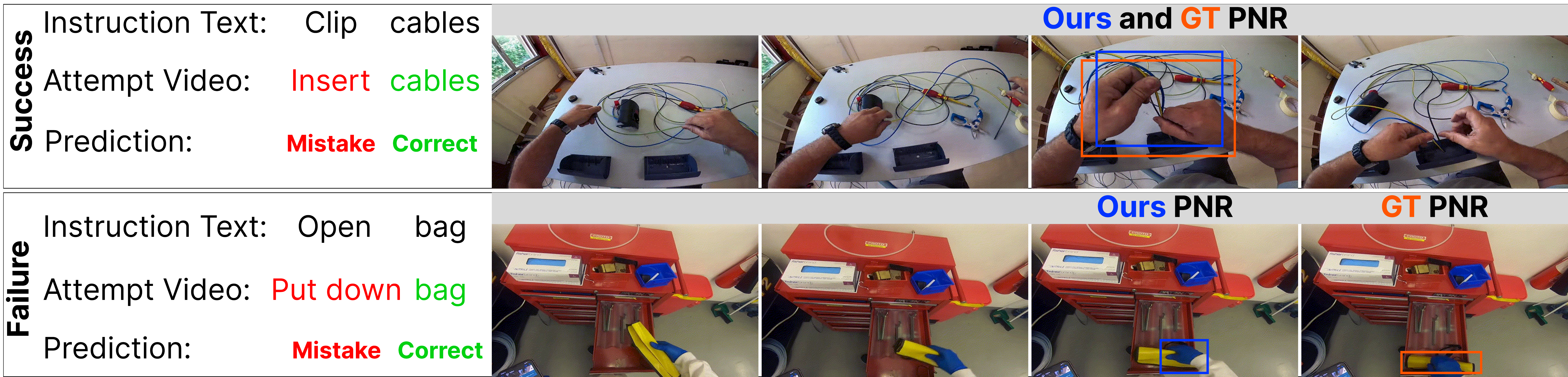}
    \caption{{Top: correct predictions (IoU=0.56). Bottom: failure case with correct semantic but incorrect temporal attribution.}}

    \label{fig:qualitative}
\end{figure}

\vspace{-8pt}
\noindent \textbf{Qualitative examples.}
\cref{fig:qualitative} shows representative outputs. Row~1 correctly produces semantic attribution, localizes the PNR frame, and generates a reasonable spatial box (IoU=0.56). Row~2 illustrates a failure case: semantic attribution is correct, but temporal attribution selects an earlier frame before the bag is set down. Judging the spatial completion state of an object from 2D video alone is inherently difficult, suggesting that depth or 3D cues are a promising direction for future work. Notably, the spatial bounding box still correctly highlights the hand performing the unintended action, which is consistent with the correct semantic prediction.

\vspace{-15pt}
\section{Implementation Details}
\vspace{-8pt}
\noindent \textbf{Training configuration.}
We train with AdamW (lr $10^{-4}$) for 20 epochs on 4$\times$H100 GPUs with batch size 64.

\end{document}